%% file: main.tex
\documentclass{article}

\usepackage[final]{neurips_2022}

\usepackage[utf8]{inputenc} 
\usepackage[T1]{fontenc}    
\usepackage{hyperref}       
\usepackage{url}            
\usepackage{booktabs}       
\usepackage{amsfonts}       
\usepackage{nicefrac}       
\usepackage{microtype}      
\usepackage{xcolor}         

\input{math_commands}

\usepackage{comment}
\usepackage{soul}
\usepackage{graphicx}
\usepackage{graphbox}
\usepackage{subfigure}

\newcommand{\norm}[2]{\| #1 \|_#2}

\usepackage{siunitx}
\DeclareFixedFont{\ttb}{T1}{txtt}{bx}{n}{12} 
\DeclareFixedFont{\ttm}{T1}{txtt}{m}{n}{12}  

\usepackage{color}
\definecolor{deepblue}{rgb}{0,0,0.5}
\definecolor{deepred}{rgb}{0.6,0,0}
\definecolor{deepgreen}{rgb}{0,0.5,0}

\usepackage{listings}

\newcommand\pythonstyle{\lstset{
language=Python,
basicstyle=\ttm,
morekeywords={self},              
keywordstyle=\ttb\color{deepblue},
emph={MyClass,__init__},          
emphstyle=\ttb\color{deepred},    
stringstyle=\color{deepgreen},
frame=tb,                         
showstringspaces=false
}}

\lstnewenvironment{python}[1][]
{
\pythonstyle
\lstset{#1}
}
{}

\newcommand\pythoninline[1]{{\pythonstyle\lstinline!#1!}}

\title{pyGSL: A Graph Structure Learning Toolkit}

\author{%
  Max Wasserman \\
  Dept. of Computer Science\\
  University of Rochester\\
  Rochester, NY 14620 \\
  \texttt{mwasser6@ur.rochester.edu} \\
  \And
  Gonzalo Mateos \\
  Dept. of Electrical and Computer Eng.\\
  University of Rochester\\
  Rochester, NY 14620 \\
  \texttt{gmateosb@ece.rochester.edu} \\
}

\begin{document}

\maketitle

\begin{abstract}
We introduce pyGSL, a Python library that provides efficient implementations of state-of-the-art graph structure learning models along with diverse datasets to evaluate them on. The implementations are written in GPU-friendly ways, allowing one to scale to much larger network tasks. A common interface is introduced for algorithm unrolling methods, unifying implementations of recent state-of-the-art techniques and allowing new methods to be quickly developed by avoiding the need to rebuild the underlying unrolling infrastructure. Implementations of differentiable graph structure learning models are written in PyTorch, allowing us to leverage the rich software ecosystem that exists e.g., around logging, hyperparameter search, and GPU-communication. This also makes it easy to incorporate these models as components in larger gradient based learning systems where differentiable estimates of graph structure may be useful, e.g. in latent graph learning. 
Diverse datasets and performance metrics allow consistent comparisons across models in this fast growing field. 
The full code repository can be found on \href{https://github.com/maxwass/pyGSL}{https://github.com/maxwass/pyGSL}.
\end{abstract}

\section{Introduction}
Graph Learning (GL), also referred to as Network Topology Inference or Graph Structure Learning, refers to the task of inferring graphs from data. Here we focus on learning undirected graphs in the supervised setting. GL has deep roots in statistics ~\citep{dempster1972covariance} with significant contributions for probabilistic graphical model selection; see e.g.~\citep{kolaczyk2009book,GLasso2008,drton2017structure}. Approaches dubbed `latent graph learning' have been used to learn interactions among coupled dynamical systems~\citep{kipf2018icml}, or to obtain better task-driven representations of relational data for machine learning (ML) applications~\citep{wang2019dynamicgraphcnn,kazi2020DGM,velickovic2020pgn}; for the related task of link-prediction see \citep{hamilton2021book} .
Graph signal processing (GSP) has been the source of recent advances using cardinal properties of network data such as smoothness~\citep{dong2016learning,kalofolias2016learn} and graph stationarity~\citep{segarra2017tsipn,pasdeloup2018tsipn}, exploiting models of network diffusion~\citep{wasserman2022GDN, daneshmand2014estimating}, or taking a signal representation approach ~\citep{dong2019learning,mateos2019spmag}. These works, referred to as Model-Based (MB) graph learning, use such data models to formulate the topology inference task as a (convex) optimization problem to be solved for each problem instance. When datasets are available, one can build on these MB approaches by unrolling an iterative solution procedure to produce a learned architecture optimized on the given data~\citep{monga2021spmag}. Such Unrolling-Based (UB) methods offer several advantages over MB methods: they tend to be more expressive (MB methods are restricted to a small set of constraints expressible in a convex manner), require fewer layers than the corresponding iterative algorithm for a given performance requirement, and allow one to directly optimize for a (differentiable) metric of interest~\citep{dong2019learning,wasserman2022GDN, shrivastava2019glad}. Thus by learning a distribution over such graphs by investing time upfront in training a model we can obtain better task-specific performance while avoiding expensively resolving each new problem instance.

While the interdisciplinary history of GL has been a source of progress, it has also been a hindrance: these different fields use different tools (MATLAB/Octave, R, Python, etc) in implementing their algorithms. This has slowed the spread of new algorithms and made comparing relative performance of different algorithms difficult. Importantly, there has been little effort to ensure implementations can leverage the increasing computation power of the GPU. Due to the reliance of GL algorithms on matrix/vector operations, as well as the interest in handling ever increasing network sizes, GPUs offer an ideal environment for such computation to occur. 
Thus the first aim of pyGSL is to provide standardized GPU-compatible implementations of fundamental GL methods, along with datasets to evaluate them on. 
The second aim is lowering the barrier to use and develop such GL methods. As such, we provide a framework which makes it easy for researchers to extend - or build completely new - graph learning methods without needing to rebuild all the underlying software machinery. By consolidating wide-ranging GL methods into in a single GPU-compatible framework, pyGSL can be seen as the most  comprehensive Python library for GL to date.

\section{Architecture}
\label{sec:architecture}

Both MB and UB methods rely heavily on matrix/vector operations, often require hyperparameter search capabilities, and are typically applied to problems with large graph sizes and/or large sets of graphs; see Section \ref{sec:gl_methods} for further details on said GL approaches. For these reasons, pyGSL implements all methods in PyTorch which has a rich ecosystem of software to interface with GPUs, run and visualize hyperparameter searches via W\&B ~\citep{wandb}, and abstract away the bug-prone training, validation and optimization logic required in gradient based training via PyTorch-Lightning ~\citep{Falcon_PyTorch_Lightning_2019}.

\begin{figure}
  \centering
  \includegraphics[scale=.3]{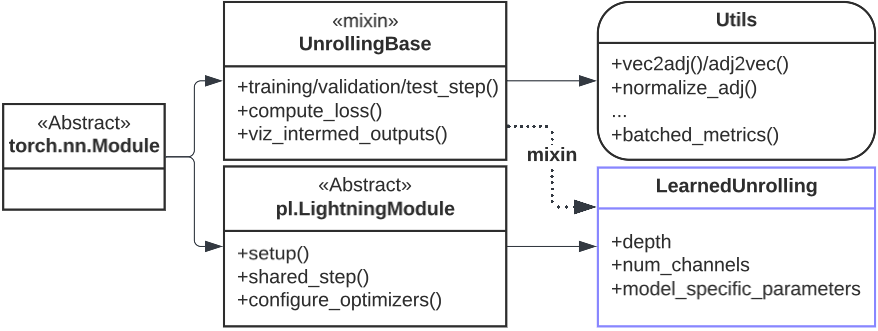}
  \caption{Class diagram of pyGSL's learned unrollings with a subset of variables and methods shown. 
  For a specific learned unrolling, say GDN, we would replace the LearnedUnrolling class with a GDN class which simply overwrites the setup(), shared\textunderscore step(), and configure\textunderscore optimizers() methods from its parent class pl.LightningModule. Since the GDN class also mixes in \texttt{UnrollingBase}, it automatically inherits functionality common to all UB methods.} 
  \label{fig:pyGSL}
\end{figure}

Implementations of UB methods share significant structure, motivating our novel \texttt{UnrollingBase} class, which encapsulates this shared functionality; see Fig. \ref{fig:pyGSL}. This minimizes repeated code, and makes implementing new UB methods as simple as defining the core layer-wise logic for that specific unrolling. By doing so, users can (i) cut their development time down significantly by relying on pretested methods; and (ii) automatically gain functionality such as intermediate output visualization and metric logging.
See Appendix \ref{app:python_example} for a code snippet showing how pyGSL can be used.

\section{A Unifying View of Graph Learning Methods}
\label{sec:gl_methods}

The GL problem is to infer a graph from data to uncover a latent complex information structure. Here, we focus on undirected and weighted graphs $\mathcal{G}(\mathcal{V},\mathcal{E})$, where $\mathcal{V}=\{1,\ldots, N\}$ is the set of nodes (henceforth common to all graphs), and $\mathcal{E}\subseteq \mathcal{V}\times \mathcal{V}$ collects the edges. A graph signal $\vx = [x_1, \ldots, x_N] \in \R^N$ is a map $\vx: \mathcal{V} \rightarrow \R$ which assigns a real value (say, a feature) to each vertex. We collect the $P$ graph signal observations together into data matrix $\mX = [\vx^{(1)}, \cdots, \vx^{(P)}]$. A similarity function $S(\mX): \R^{N \times P} \mapsto \R^{N \times N}$ is chosen to compute the observed direct similarity between nodes. Common choices for $S$ include sample covariance/correlation or Euclidean distance. 
We can conceptualize the output of $S(\mX)$ as the symmetric adjacency matrix of an observed graph 
from which we would like to recover a latent graph with symmetric adjacency matrix denoted $\mA_L \in \mathbb{R}_+^{N\times N}$. Below we discuss three popular approaches to tackle the GL task which exemplify the link between $S(\mX)$ and $\mA_L$; for discussions on Deep Learning (DL) methods for GL and tradeoffs between the three methods refer to \ref{app:DL} and \ref{app:tradeoffs}, respectively.

%
\begin{table}[t]
\caption{Model-Based methods with the associated Unrolling-Based method it inspires.}
\label{table:model_based_methods}
\begin{center}
\begin{tabular}{lll}
\multicolumn{1}{c}{\bf Data Model} 
&\multicolumn{1}{c}{\bf MB Iterative Solution Procedure}
&\multicolumn{1}{c}{\bf UB Model}
\\ \hline \\
Gaussian 
& Alternating Minimization
& GLAD\\
Smoothness       
& Primal-Dual Splitting
& L2G\\
Diffusion 
& Proximal Gradient Descent
& GDN
\end{tabular}
\end{center}
\end{table}

\textbf{Model-Based Graph Learning.}
These GL approaches postulate some data model relating the observed data $\mX$ to the latent graph $\mathcal{G}_L$ via $\mX \sim \mathcal{F}(\mA_L)$. 
This model can be the result of a network process, e.g. linear network diffusion, or statistical where $\mX$ follows a distribution determined by $\mA_L$ 
e.g. in probabilistic graphical models. Thus the GL task reduces to an attempt to invert this relation $\mathcal{F}^{-1}(\mX)$. 
To do so, these techniques formulate a (convex) optimization problem which can be solved via iterative optimization methods. 
These MB GL problems have the general form
\begin{align}
\label{prob:model-based-graph-learning}
\mA^{*} \in {}& \underset{ \mA \in \mathcal{C} }  {\text{ argmin}} \: \{\mathcal{L}_{\text{data}}(\mA,\mX) + 
\mathcal{L}_{\text{reg}}(\mA)\},
\end{align}
where $\mathcal{L}_{\text{data}}(\mA,\mX)$ is the data fidelity term, $\mathcal{L}_{\text{reg}}(\mA)$ is the regularization term incorporating the structural priors (e.g., $\norm{\mA}{1}$ for sparsity), and $\mathcal{C}$ encodes a convex constraint on the optimization variable $\mA$, e.g., symmetry, non-negativity, or hollow diagonal. We inject our assumptions about the generative model into the objective of the problem. Note that we parameterize the canonical problem form with an adjacency $\mA$ for notational convenience, but many methods use other graph shift operators such as the Laplacian $\mL := \text{diag}(\mA \mathbf{1}) - \mA$ or its normalized counterparts.

To make this clear, consider the following data models and resulting optimization formulations for graph recovery, summarized in Table \ref{table:model_based_methods}. When the data $\mX$ are assumed to be Gaussian, the GL task is to estimate the precision matrix; the sparsity-regularized log-likelihood function for such a precision matrix is convex 
$\arg\min_{\mathbf{A}\succeq\mathbf{0}}\left\{{\textrm{Tr}(S(\mX) \mathbf{A})} {+\alpha\|\mathbf{A}\|_1 - \beta\log\det\mathbf{A}} \right\}$, where $S(\mX)$ is the sample covariance matrix~\citep{GLasso2008}.
When the data is assumed to be smooth on $\mathcal{G}_L$, i.e., the total variation Tr$(\mX^\top \mL \mX)$ of the signals on the graph is small, a standard GL formulation is $\arg\min_{\mathbf{A} \in \mathcal{C}} \left\{ {\norm{\mA \odot S(\mX)}{1}} {- \alpha \mathbf{1}^\top \text{log}(\mA \mathbf{1}) + \beta \norm{\mA}{F}^2} \right\}$, where $S(\mX)$ is the Euclidean distance matrix~\citep{kalofolias2016learn}. 
When the observed graph $S(\mX)$ satisfies a graph convolutional relationship with $\mA_L$, i.e. 
$S(\mX) = \sum_{i=0}^{K} \alpha_i \mA_L^i$, as is the case in linear network diffusion where again $S(\mX)$ is the sample covariance matrix, we can pose the GL task as the non-convex problem
$\arg\min_{\mathbf{A} \in \mathcal{C}} \left\{ {\norm{S(\mX)-\sum_{i=0}^{K} \alpha_i \mA_L^i}{F}^2} {+ \beta\norm{\mA}{1}} \right\}$~\citep{wasserman2022GDN}.

The MB methods that admit iterative solutions take the generic form: $\mA[i+1] = h_{\mathbf{\theta}}(\mA[i], S(\mX))$ where $\mA[i]$ is output on the $i$-th iteration, $h_{\mathbf{\theta}}$ is the contractive function, and $\mathbf{\theta}$ are the regularization parameters. We implement the function $h_{\mathbf{\theta}}$ using GPU compatible operations in PyTorch and wrap it in a loop until uniform convergence is achieved. 

\begin{figure}[t]
\centering
  \begin{minipage}[b]{0.49\textwidth}
    \includegraphics[width=\textwidth]{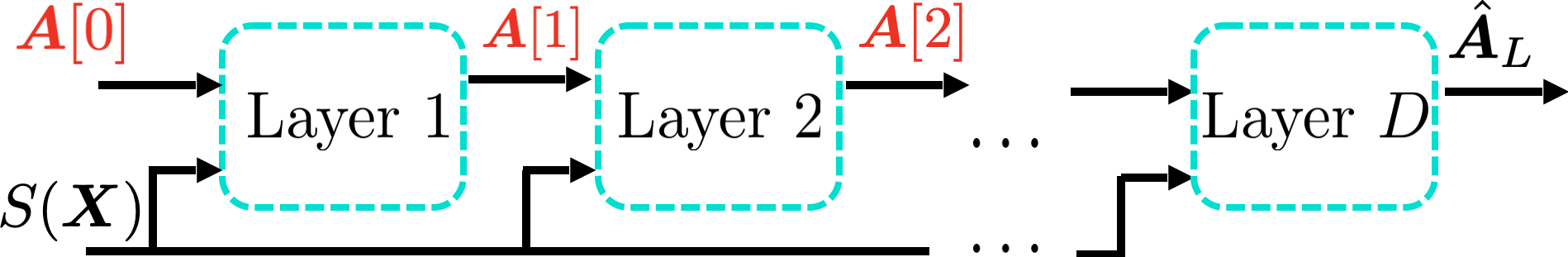}
    \caption{{Schematic of a generic UB method.}}
    \label{fig:generic_UB}
  \end{minipage}
  \begin{minipage}[b]{0.49\textwidth}
    \includegraphics[width=\textwidth]{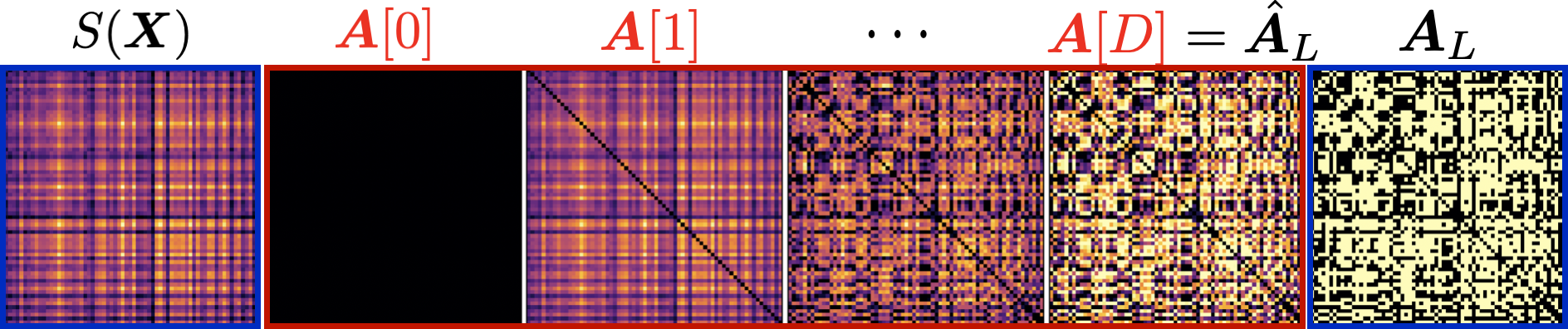}
    \caption{{Intermediate outputs plotted via pyGSL.}}
    \label{fig:iterative_optim}
  \end{minipage}
\end{figure}

\textbf{Unrolling-Based Graph Learning.}
Algorithm unrolling uses iterative algorithms, like those often used in signal processing, as an inductive bias in the design of the neural network architectures. The algorithm is unrolled into a deep network by associating each layer to a single iteration in the truncated algorithm, producing a finite number of stacked layers as shown in Figure \ref{fig:generic_UB}. We transform the regularization parameters of the iterative algorithms into learnable parameters of the neural network. Using a dataset, such parameters can be optimized for a given task by choosing an appropriate loss function and backproprogating gradients; see ~\citep{monga2021spmag} for more details.

All UB models share a significant amount of code in their construction, training, and evaluation. We consolidate this code into the novel \texttt{UnrollingBase} class. To instantiate a new UB model, one simply inherits from \texttt{UnrollingBase} and Pytorch-Lightning's \texttt{LightningModule}, declares a layers' learnable parameters $\theta^{i}$, and implements the $h_{\mathbf{\theta}^i}(\mA[i], S(\mX))$ 
function - denoted as \texttt{shared\_step} in the users class. 
Recall that each layer of the resulting unrolled network performs the same operation represented by $h_{\mathbf{\theta}}(\mA[i], S(\mX))$ here; thus specifiying the entire unrolled network reduces to providing a desired depth and specifying $h$ for a single layer.
The \texttt{UnrollingBase} mixin takes care of stacking the layers together, feeding outputs of one layer as inputs of the next, and implements the required interface to allow the \texttt{LightningModule} to automate the training and optimization.

\section{Data}
\label{sec:data}
\textbf{Synthetics}
As discussed in Section \ref{sec:gl_methods}, MB methods motivate the optimization problem through an assumed generative model on the observed data. Since UB methods are inspired by their MB counterparts, 
synthetic datasets which match these assumed generative models are important for model validation across GL methods.
We thus provide (i) an efficient graph sampling interface using NetworkX ~\citep{networkx} for a broad range of random graph ensembles;
and (ii) an ability to generate signals conforming to 
smoothness, 
linear network diffusion ($\vx = \sum_{i=0}^p \alpha_i A_L^{i}\vw$, where $\vw$ is typically white), or 
Gaussianity ($\vx \sim \mathcal{N}(\mathbf{0}, \mA^{-1}$)). These datasets can be accessed via the \texttt{smooth}, \texttt{diffuse}, and \texttt{gaussian} classes, respectively, in the \texttt{gl.data} subdirectory.

\textbf{Real Data} We provide real datasets to evaluate the models on, including over $1000$ Structural Connectivity (SC) and Functional Connectivity (FC) pairs extracted from the HCP-YA neuroimaging dataset ~\citep{glasser2016human}, as well as co-location and social network data
from the Thiers13 dataset ~\citep{Genois2018}. We also provide the ability to create `Pseudo-synthetic' data by using the graphs from the real datasets, and sampling synthetic signals on top of them that are e.g., smooth, diffused, or Gaussian. This provides a gentler transition to real data, which we found useful when developing and validating GL methods.

See Appendix \ref{app:data} for a further discussion on the data and its associated class layout in pyGSL. 

\section{Related Work}
While there has been an explosion of software development in adjacent fields such as geometric DL~\citep{pytorch_geom} and network-based modeling of complex systems~\citep{deepgraph}, there has been little in the way of comprehensive GL packages. 
\href{https://epfl-lts2.github.io/gspbox-html/doc/learn_graph/}{GSPBox} is (unmaintained) Matlab toolkit which focuses on traditional GSP tasks and offers a single function to perform GL on the smooth signal case ~\citep{perraudin2016gspbox}.
There are many algorithms for Gaussian graphical model selection from a likelihood based ~\citep{Yuan2007biometrika, GLasso2008}, regression based ~\citep{meinshausen2006high, peng2009partial}, and constrained 
$\ell 1$-minimization
~\citep{cai2011constrained} approaches; almost all implementations are in R or Matlab. Recently~\citet{choi2021efficient} released an R-based GPU implementation of CONCORD-PCD, a regression based approach which uses parallelized coordinate descent for fast inference. Software in~\citep{lassance2020mlsp} offers benchmarks for MB approaches.
To the best of the authors knowledge, no encompassing framework exists for UB methods, only standalone implementations released by the respective methods' authors.

\section{Concluding remarks}
We introduced the pyGSL framework for GPU-aware fast and scalable GL. We provide several synthetic and real datasets for model evaluation, as well as a novel class for UB architectures which dramatically lowers the time and expertise required to use and build such methods. It also ensures their compatibility as sub-modules in larger gradient-based learning systems. This is an active project, and we plan to add more benchmark datasets and the latest GL methods as they are published. We welcome more researchers and engineers to join, develop, maintain, and improve this toolkit to push forward the research and deployment of network topology inference algorithms.

\bibliographystyle{unsrtnat}
\bibliography{main}

\appendix

\newpage
\section{Appendix}

\subsection{Data Class Diagram}
\label{app:data}
\begin{figure} 
  \label{fig:data_class}
  \centering
  \includegraphics[scale=.25]{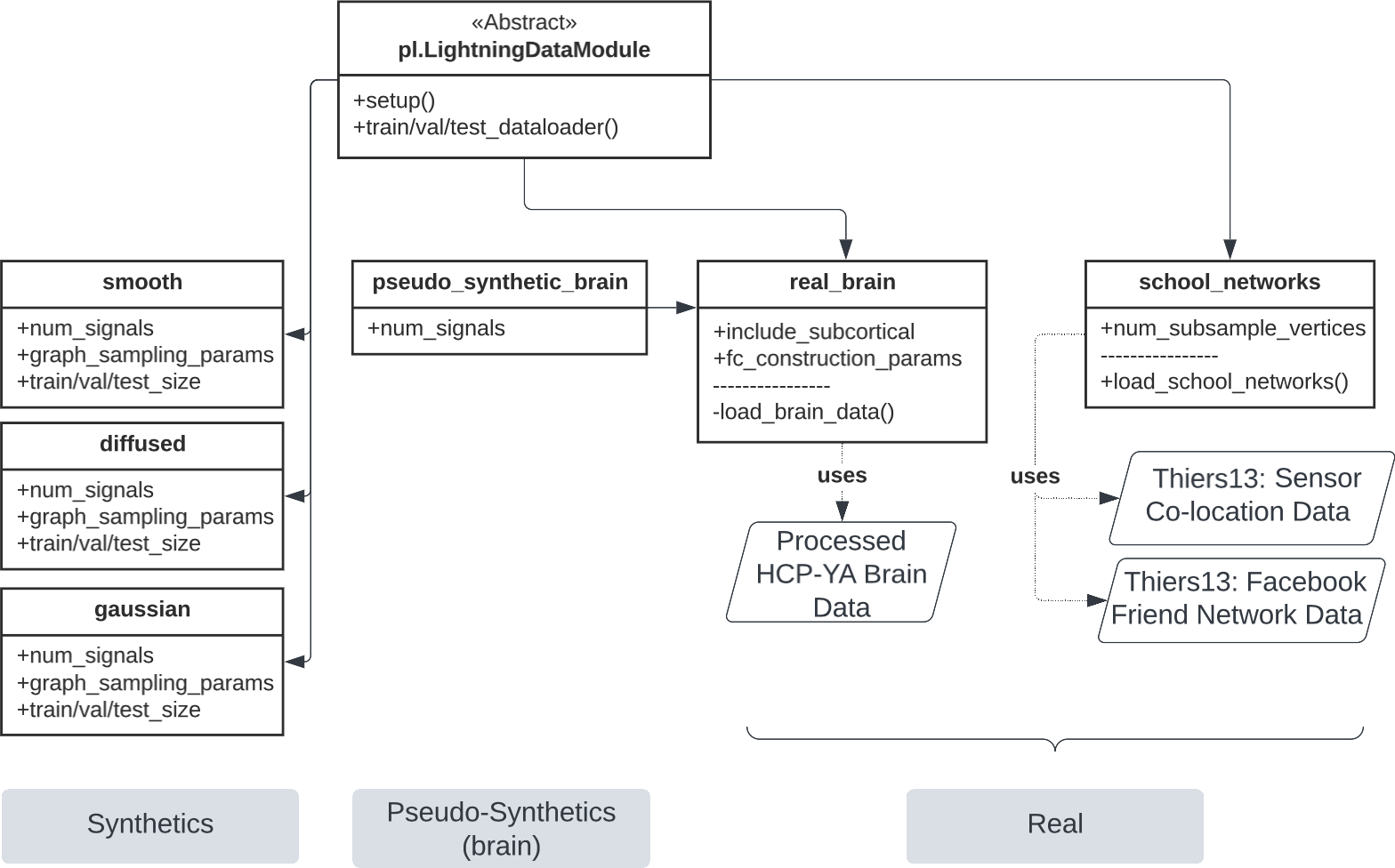}
  \caption{Class diagram of pyGSL's provided datasets}
\end{figure}

There are three main categories of data provided by pyGSL as shown in \ref{fig:data_class}. The first being Synthetics, which are constructed as to be faithful to a generative model assumed by at least one of the provided MB methods, e.g. diffused and GDNs. The second are Real Data, particularly FC-SC pairs derived from HCP-YA neuroimaging data ~\citep{van2013wu} and a large Sensor Co-location - Facebook Friend Network graph pair derived from the Theirs13 dataset ~\citep{Genois2018}.
The last category of data are the `Pseudo-synthetic' data which is a compromise between real and synthetic data: we use real observed graphs but sample synthetic signals - either smooth, diffused, or Gaussian - on top of them. This provides a smoother transition from synthetic to the more difficult real data.

Raw HCP-YA data can be found \hyperlink{https://www.humanconnectome.org/study/hcp-young-adult/overview}{here}, and use of raw data - and derivative forms of such data - is subject to the \hyperlink{http://www.humanconnectomeproject.org/wp-content/uploads/2010/01/HCP_Data_Agreement.pdf}{HCP Data Use Agreement}. If the processed SC's are used, please cite~\citep{zhengwu2018-SC-population, zhengwu219-SC-tensor}.

\subsection{Deep Learning Approach to GL}
\label{app:DL}
Classic Graph Neural Networks (GNN) can be manipulated to perform the GL task, often by decoding latent node feature vectors into predicted edges. Due to low inductive bias and poor scaling properties DL models struggle on many GL tasks, but we include two popular GNNs, Graph Convolutional Network (GCN) ~\citep{kipf2017iclr} and Graph Isomorphism Networks (GIN) ~\citep{XuGIN}, with appropriate adjustments for tackling the GL task.

\subsection{Tradeoffs in GL Approaches}
\label{app:tradeoffs}
Factors which influence the decision of which GL method to use include training and inference complexity, scaling properties, expressive power, parameter efficiency, and empirical performance.

Inference in MB methods involves solving an optimization problem which often has cubic run-time and memory complexity w.r.t. the number of nodes $\mathcal{O}(N^3)$. Training in MB methods involves performing a hyperparameter search, which involves solving many such optimization problems, posing the same problematic run-time and memory issues. MB methods are highly parameter (here regularization coefficients) efficient - a characteristic of convex optimization problems, but lack expressivity due to the limited constraints expressible in a convex form. These methods are empirically observed to perform well when the graphs are small enough such that comprehensive searches over the regularization coefficients can be performed.

Inference and training in DL methods involves forward passes (and backward passes for training) through the deep network which is typically quadratic in run-time and memory w.r.t the number of nodes $\mathcal{O}(N^2)$. These methods tend to be highly expressive, and while more parameter efficient than MLPs due to parameter sharing, they still typically require on the order of $10^3 - 10^6$ parameters, and often even more, depending on the particular GNN. For reasonable performance they tend to require a significant amount of data, more than is typically available in most use cases. 

While one could categorize UB methods as DL methods - because they do in fact generate a deep NN architecture - they are unique enough for their own category. Training and inference is identical to other DL methods, involving forward passes of the data through the network (and the corresponding backward pass to propagate gradients for training) with the associated quadratic run-time and memory complexities. These methods have significant inductive bias inherited from the associated MB iterative algorithm, making them highly parameter efficient but also more expressive because the (now learnt) parameters can be tuned on a dataset via gradient based learning. Empirically, these methods tend to perform best in most cases.

An additional note is that MB and UB methods are naturally inductive, meaning you can train on graphs of size $N_{train}$ and deploy that same model on graphs of size $N_{test} >> N_{train}$, while - in general - DL methods do not have this property. This is because DL methods can often use features of the graph which functionally depend on the graph size - tieing them to the particular graph size in inference. With some adjustment, it is possible to break this dependence and make these methods inductive.

\subsection{Python Example}
\label{app:python_example}
\begin{figure}[!htb]
\label{fig:python_example}
\begin{python}
import graph_learning as gl
import pytorch-lightning as pl

graph_sampling = {'graphs': 'ER', 'n': 100, 'p': 0.5}
data_size = {'train': 500, 'val': 100, 'test': 100}
dm = gl.data.smooth(graph_sampling, **data_size)

model = gl.models.unroll.glad(depth=20)

trainer = pl.Trainer(max_epochs=200, 'gpus': 1)
trainer.fit(model, datamodule=dm)
trainer.test(model, datamodule=dm, ckpt_path='best')
\end{python}
\caption{Example of how to use the package. All training loops, optimization, and GPU communication are abstracted away, simplifying the interface.}
\label{fig:sample_code}
\end{figure}

Figure \ref{fig:sample_code} shows how pyGSL would be used in a simple example of training a GLAD network - an UB method motived from a MB method which postulaes a Gaussian generative data model - of depth $20$ on 'smooth' data, see Section \ref{sec:gl_methods} for details.

\end{document}

%% file: math_commands.tex
\usepackage{amsmath,amsfonts,bm}

\def\eqref#1{equation~\ref{#1}}

\def\1{\bm{1}}

\def\vw{{\bm{w}}}
\def\vx{{\bm{x}}}

\def\mA{{\bm{A}}}

\def\mL{{\bm{L}}}

\def\mX{{\bm{X}}}

\DeclareMathAlphabet{\mathsfit}{\encodingdefault}{\sfdefault}{m}{sl}
\SetMathAlphabet{\mathsfit}{bold}{\encodingdefault}{\sfdefault}{bx}{n}

\newcommand{\R}{\mathbb{R}}

